\documentclass[11pt]{article}

\usepackage[utf8]{inputenc}
\usepackage[T1]{fontenc}
\usepackage{lmodern}
\usepackage[margin=1in]{geometry}
\usepackage{amsmath,amssymb}
\usepackage{graphicx}
\usepackage{booktabs}
\usepackage{multirow}
\usepackage{array}
\usepackage{float}
\usepackage{caption}
\usepackage[numbers,sort&compress]{natbib}
\usepackage{xcolor}
\usepackage[colorlinks=true,linkcolor=blue!60!black,citecolor=blue!60!black,urlcolor=blue!60!black]{hyperref}
\usepackage{url}
\usepackage{authblk}
\usepackage{orcidlink}

\title{Dataset-Specific Encoding of Image Content Underlies Cross-Dataset Failure in Brain MRI Foundation Models}

\author[1]{Minh Sao Khue Luu\,\orcidlink{0000-0002-6766-4966}}
\author[1]{Bair N. Tuchinov\,\orcidlink{0000-0002-8931-9848}}
\affil[1]{The Artificial Intelligence Research Center of Novosibirsk State University, Novosibirsk 630090, Russia}
\affil[ ]{\texttt{luumsk@gmail.com}}

\date{}

\begin{document}

\maketitle

\begin{abstract}
Frozen features of brain MRI foundation models lose accuracy on new datasets,
a failure usually blamed on the features encoding where each scan came from.
We show that the failure lies elsewhere. Across 17 public datasets and eight
frozen representations, every representation recognizes the MRI sequence
within each dataset, yet most fail on an unseen one. The cause is neither
dataset encoding nor a shift of each dataset: the content itself is encoded
along different directions in each dataset. This explains which datasets
fail, for the sequence and for sex, and why removing the dataset signal with
INLP, LEACE, CORAL or adversarial training never improves sequence transfer.
Only the broadly pretrained 3DINO and SAM-Med3D encode the sequence
consistently enough to transfer; the brain-specific BrainIAC and BrainFM do
no better than an untrained network.
Code is available at
\url{https://github.com/BrainFM/brainfm-profiler/tree/main/representation_utility}.
\end{abstract}

\section{Introduction}

Public brain MRI datasets support machine learning research on neurological
disease but are individually small and narrow. Foundation models for brain
MRI are therefore pretrained on pooled collections drawn largely from public
datasets~\cite{munk_large-scale_2025, tak_generalizable_2026,
liu_modality-agnostic_2025}. Pooling increases sample size but introduces
heterogeneity. Dataset bias is well known from natural
images~\cite{torralba_unbiased_2011}, and brain MRI adds to it: each public
dataset mostly covers one disease, cohort and age range, so dataset identity
is confounded with clinical content; sequences differ in contrast;
release-specific preprocessing cannot be undone; labels overlap little across
datasets; and subjects may recur across releases~\cite{jimenez-sanchez_copycats_2024}.
Models trained on such pools may learn dataset shortcuts~\cite{geirhos_shortcut_2020,
zech_variable_2018} rather than anatomy or pathology.

It is well established that representations retain where a scan came from.
The source study or site can be decoded from brain
MRI~\cite{wachinger_detect_2021, glocker_machine_2019}, and recent negative
controls show that this signal is already present in randomly initialized
encoders and in the raw image~\cite{rahbar_frozen_2026}, with the same pattern
in EEG and fMRI foundation models~\cite{zare_negative-control_2026,
tao_batch_2026}. Frozen medical foundation models also lose accuracy on
datasets and centres not seen in training, as shown for
mammography~\cite{germani_multi-institutional_2026, nguyen_benchmarking_2026}
and pathology~\cite{de_jong_current_2025, komen_towards_2025}. That frozen
features fail across datasets is therefore known; why they fail is not.
A representation can encode dataset identity and still transfer content,
provided the content is encoded in the same way across datasets, so dataset
identity alone does not explain the failure. For brain MRI, existing tests
also do not cover the setting in which public data are actually pooled. They
move between sites of a single study~\cite{rahbar_frozen_2026} or between
scanners imaging the same subjects~\cite{navarro-gonzalez_cross-scanner_2026},
or they apply simulated artifacts~\cite{mielcarz_3d_2026}. In the only brain
MRI test of pretrained models against an untrained control, pretraining gave
no reliable advantage~\cite{rahbar_frozen_2026}, but only models of one
architecture family were tested. Removing the site signal after extraction
did not help on held-out sites~\cite{rahbar_frozen_2026}, but whether this
holds for other removal methods and for independent datasets, and what
prevents it from working when the content is clearly encoded, has not been
measured.

We address these questions with frozen representations, the way foundation
models are commonly used. Transfer is measured by leave-one-dataset-out
linear probing on 17 independent public datasets: a linear classifier trained
on the frozen features of all datasets but one is evaluated on the held-out
dataset. The probe predicts the MRI sequence (T1, T1c, T2 or FLAIR), the only
content label available for every scan in all 17 datasets; disease labels
barely overlap between datasets and are confounded with dataset identity. Sex,
available in ten datasets, serves as a second, anatomical label. Because the sequence is easily
recognized within every dataset, a failure to transfer it cannot be blamed
on missing information, which lets us study why transfer fails rather than
only whether it does. In short, we find that in most representations the
difference between sequences is itself different in each dataset: this is
not a shift, so centring does not repair it, and it ties the dataset signal
to the content, so removing one removes the other. Our contributions are:

\begin{itemize}
  \item \textbf{A diagnosis of cross-dataset failure that also explains why
  removing the dataset signal does not help.} We separate three candidate
  causes of the failure: encoding of the source dataset, a shift of each
  dataset, and disagreement of the class directions between datasets. To tell
  them apart, we combine centring tests with a direction-consistency score
  adapted from cross-lingual transfer~\cite{muller_first_2021,
  deshpande_when_2022}. With these tools we identify the cause for the
  sequence and for sex, rank representations from the training datasets
  alone, and explain why four removal methods (INLP, LEACE, CORAL and
  adversarial training) fail to improve transfer, revising the earlier
  explanation of this failure on held-out sites~\cite{rahbar_frozen_2026}.
  \item \textbf{A controlled comparison of brain-specific and broadly
  pretrained foundation models.} We compare two brain-specific and two
  broadly pretrained foundation models against four baseline encoders, with
  controls for architecture, input scaling, skull stripping and pretraining
  overlap. We show that broad multi-organ pretraining, not brain-specific
  pretraining, yields features that transfer across datasets, qualifying the
  earlier finding that pretraining adds nothing~\cite{rahbar_frozen_2026}.
\end{itemize}

\noindent We release the code, the list of sampled scans and the result
tables.

\section{Related Work}

\paragraph{Dataset and site signatures.}
``Name that dataset'' experiments showed that natural-image datasets are
easily told apart~\cite{torralba_unbiased_2011}. In brain MRI, the source
study can be predicted from scans of 17 studies~\cite{wachinger_detect_2021}
and remains predictable after standard
preprocessing~\cite{glocker_machine_2019}. In medical imaging more broadly,
models can exploit site information as a shortcut that fails at new
hospitals~\cite{zech_variable_2018, geirhos_shortcut_2020}. Most closely
related, Rahbar~\cite{rahbar_frozen_2026} probes three SwinUNETR encoders
(brain-pretrained, CT-pretrained and untrained) on the multi-site ABIDE
cohorts. Site is decodable at a similar level from all three and from the raw
image, and pretraining shows no reliable advantage for any clinical target.
Removing the site directions with INLP or ComBat gives no significant gain on
held-out sites, which Rahbar attributes to the averaged encoder output carrying little
clinical information to begin with, and it damages segmentation, because the
directions that carry site also carry anatomy. Similar negative controls show that EEG foundation
models decode dataset identity perfectly~\cite{zare_negative-control_2026},
and fMRI foundation-model embeddings are dominated by batch
effects~\cite{tao_batch_2026}. Our work differs in four ways: the domains are
17 independent public datasets rather than sites of one study; the outcome is
the transfer of a content label that every representation encodes well within
each dataset; the representations include broadly pretrained foundation
models, for which we do find a pretraining advantage; and we measure why
transfer and removal fail.

\paragraph{Cross-dataset failure of frozen medical foundation models.}
Outside brain MRI, several studies show that frozen foundation-model features
fail across datasets or centres. Germani et
al.~\cite{germani_multi-institutional_2026} use frozen encoders on ten
mammography datasets and find that classifiers trained on the aggregated data
show dataset-dependent failure modes, with disparities in error, confidence
and calibration, and they propose representation-level diagnostics of
shortcut risk and evaluate mitigation strategies. Nguyen et
al.~\cite{nguyen_benchmarking_2026} probe 15 frozen backbones trained on three
mammography datasets and tested on twelve others: mammography-specific
pretraining does not guarantee robustness, a general-purpose vision model
remains competitive, and the better representations keep clinical signal while
still reflecting dataset and acquisition structure. In pathology, all ten
foundation models tested by de Jong et al.~\cite{de_jong_current_2025}
represent the medical centre strongly, and errors follow same-centre
confounders; PathoROB~\cite{komen_towards_2025} extends this to 20 models and
34 centres, and image-level processing does not remove the hospital
signatures~\cite{komen_histopathological_2024}. These studies establish that
the failure occurs, and Germani et al.\ add diagnostics of shortcut risk at
the level of the representation. Our aim is to explain the failure in brain
MRI, by measuring whether the directions that separate a content label
agree across datasets, by separating a shift of each dataset from a
disagreement of directions, and by showing why removing the dataset signal
cannot repair the second.

\paragraph{Robustness of brain MRI foundation models.}
Recent 3D foundation models include brain-specific models such as
BrainIAC~\cite{tak_generalizable_2026} and
BrainFM~\cite{liu_modality-agnostic_2025}, and broadly pretrained medical
models such as 3DINO~\cite{xu_generalizable_2025} and
SAM-Med3D~\cite{wang_sam-med3d_2025}. Mielcarz et al.~\cite{mielcarz_3d_2026}
apply seven simulated MRI artifacts to five such models and find 3DINO the
most stable and BrainIAC the most sensitive, but do not test transfer across
real datasets. Navarro-Gonz\'alez et
al.~\cite{navarro-gonzalez_cross-scanner_2026} scan the same 20 subjects on
eight scanners and find that encoders supervised with anatomical or age
metadata are scanner-robust, whereas self-supervised ones, including BrainIAC,
are not. Mouheb et al.~\cite{mouheb_medical_2026} compare BrainIAC and 3DINO
on African and non-African cohorts and relate performance gaps to dataset
size. Avci et al.~\cite{avci_metadata_2026} learn representations that
separate anatomy from acquisition by modelling acquisition metadata. Our
results agree with the artifact study on real dataset shift, and our
controls attribute the difference between models to pretraining rather than
to architecture, input scaling or data overlap.

\paragraph{Removing unwanted information from representations.}
Feature-level harmonization of MRI~\cite{yang_harmonization_2025}, such as
ComBat~\cite{johnson_adjusting_2007, fortin_harmonization_2018}, adjusts
features for site effects. In representation learning, iterative nullspace
projection (INLP)~\cite{ravfogel_null_2020} and LEACE~\cite{belrose_leace_2023}
erase a labelled attribute linearly; CORAL~\cite{sun_deep_2016} aligns
second-order statistics between domains; and domain-adversarial
training~\cite{ganin_domain-adversarial_2016}, also used to unlearn scanner
information in neuroimaging~\cite{dinsdale_deep_2021}, makes a domain
classifier fail. FEATMAP~\cite{donle_featmap_2026} corrects acquisition
signatures in foundation model embeddings using paired scans of the same
subjects. We test whether such methods help when only a dataset label is
available, as is the case for public collections.

\paragraph{Predicting transfer from representation geometry.}
In multilingual language models, the alignment of representations across
languages predicts zero-shot cross-lingual transfer~\cite{muller_first_2021,
deshpande_when_2022}. In domain generalization, aligning class-conditional
features across source domains is a common training objective, although it
is not sufficient for generalization on its own~\cite{mahajan_domain_2021},
and geometric properties of in-distribution embeddings have been used to
predict robustness to corruptions without target
labels~\cite{zia_representation_2026}. We bring the alignment idea to frozen
medical encoders as a diagnostic: the agreement of per-dataset sequence
directions, computed from class means, and a centring test that separates a
shift of each dataset from a disagreement of directions. Leave-one-domain-out
evaluation follows standard practice in domain
generalization~\cite{gulrajani_search_2021}, and linear probes on frozen
features follow~\cite{alain_understanding_2017}.

\section{Data and Representations}
\label{sec:data}

\subsection{Datasets}

We use 17 public adult brain MRI datasets that cover tumors, multiple
sclerosis, stroke, epilepsy, neurodegeneration and healthy controls
(Table~\ref{tab:datasets}). They were selected from a survey of public brain
MRI datasets~\cite{luu_structured_2025}, keeping those with well-formed NIfTI
files and replacing aggregated benchmarks by their original sources. Each
dataset contributes at most about 300 structural volumes, sampled
patient-wise (all scans of a randomly chosen patient) until the cap is
reached, so that no dataset dominates; this yields 4{,}167 volumes. Ten
datasets contain more than one sequence and seven contain only one. A
screen of cross-dataset nearest neighbours in four representations, checked
against acquisition metadata, found no evidence of subjects shared between
datasets; the only matches were among the BraTS25 sub-challenges, which share
a processing grid, not subjects.

\begin{table}[t]
\centering
\small
\caption{The 17 public brain MRI datasets, grouped by domain. \emph{Sampled}:
number of volumes used in this study.}
\label{tab:datasets}
\begin{tabular}{llr}
\toprule
\textbf{Dataset} & \textbf{Sequences} & \textbf{Sampled} \\
\midrule
\multicolumn{3}{l}{\textit{Brain tumor}} \\
BraTS25-MEN~\cite{labella_asnr-miccai_2023, labella_analysis_2024, labella_multi-institutional_2024} & T1, T1c, T2, FLAIR & 300 \\
BraTS25-MET~\cite{maleki_analysis_2025}                          & T1, T1c, T2, FLAIR & 300 \\
BraTS25-SSA~\cite{adewole_brain_2023, adewole_brats-africa_2025} & T1, T1c, T2, FLAIR & 300 \\
BrainMetShare~\cite{grovik_brainmetshare_2020}                   & T1, T1c, FLAIR & 300 \\
NSK-GBM~\cite{filimonova_glioblastoma_2026}                       & T1, T1c, T2, FLAIR & 168 \\
UPENN-GBM~\cite{bakas_multi-parametric_2021, bakas_university_2022} & T1, T1c, T2, FLAIR & 300 \\
\midrule
\multicolumn{3}{l}{\textit{Multiple sclerosis}} \\
3D-MR-MS~\cite{lesjak_novel_2018}                                & T1, T1c, T2, FLAIR & 120 \\
MS-60~\cite{ali_m_muslim_brain_2022}                             & T1, T2, FLAIR & 180 \\
MSLesSeg~\cite{guarnera_mslesseg_2025}                           & T1, T2, FLAIR & 300 \\
\midrule
\multicolumn{3}{l}{\textit{Stroke}} \\
ISLES22~\cite{hernandez_petzsche_isles_2022}                     & FLAIR & 250 \\
\midrule
\multicolumn{3}{l}{\textit{Epilepsy}} \\
EPISURG~\cite{perez-garcia_episurg_2020}                         & T1 & 300 \\
\midrule
\multicolumn{3}{l}{\textit{Neurodegenerative}} \\
OASIS-1~\cite{marcus_open_2007}                                  & T1 & 300 \\
OASIS-2~\cite{marcus_open_2010}                                  & T1 & 301 \\
UMF-PD~\cite{badea_exploring_2017}                               & T1 & 82 \\
\midrule
\multicolumn{3}{l}{\textit{Healthy}} \\
BGSP~\cite{buckner_brain_2014}                                   & T1 & 241 \\
IXI~\cite{noauthor_ixi_nodate}                                   & T1, T2, PD & 300 \\
NFBS~\cite{puccio_preprocessed_2016}                             & T1 & 125 \\
\midrule
\textbf{Total}                                                   & & 4{,}167 \\
\bottomrule
\end{tabular}
\end{table}

\subsection{Representations}

A \emph{representation} is the feature vector obtained for each volume from a
frozen encoder together with its input preprocessing; no model is
fine-tuned. We compare eight representations in three groups
(Table~\ref{tab:representations}): four baseline encoders, ranging from no
learning to CT pretraining, with the three networks implemented in
MONAI~\cite{cardoso_monai_2022}; two brain-specific foundation models; and
two broadly pretrained medical foundation models. The Handcrafted
representation is a 26-element vector of intensity percentiles and moments,
entropy, gradient and sharpness measures, and the original voxel geometry.
The feature choice (class token or pooled map) and intensity scaling for
BrainIAC, BrainFM and 3DINO follow Mielcarz et al.~\cite{mielcarz_3d_2026},
within the shared geometry described below; for SAM-Med3D, the image-encoder output is
mean-pooled.

Three control representations aid interpretation: the ViT-L of 3DINO with
random weights; 3DINO with the shared $z$-score input instead of its
percentile scaling; and MedicalNet with percentile scaling instead of its
$z$-score input.

\begin{table}[t]
  \centering
  \small
  \caption{Frozen representations compared in this study. Dim.: feature
  dimension.}
  \label{tab:representations}
  \begin{tabular}{llp{5.6cm}r}
    \toprule
    Representation & Architecture & Pretraining & Dim. \\
    \midrule
    \multicolumn{4}{l}{\emph{Baseline encoders}} \\
    Handcrafted & --- & none (image statistics) & 26 \\
    Untrained CNN & DenseNet-121~\cite{huang_densely_2017} & none (random weights) & 1024 \\
    MedicalNet~\cite{chen_med3d_2019} & ResNet-34~\cite{he_deep_2016} & supervised segmentation, MRI and CT & 512 \\
    SwinUNETR~\cite{hatamizadeh_swin_2022-1} & Swin & self-supervised, CT~\cite{tang_self-supervised_2022} & 768 \\
    \midrule
    \multicolumn{4}{l}{\emph{Brain-specific foundation models}} \\
    BrainIAC~\cite{tak_generalizable_2026} & ViT-B/16 & self-supervised, brain MRI & 768 \\
    BrainFM~\cite{liu_modality-agnostic_2025} & 3D U-Net & brain MRI with synthetic contrast, resolution and bias-field changes & 2048 \\
    \midrule
    \multicolumn{4}{l}{\emph{Broadly pretrained medical foundation models}} \\
    3DINO~\cite{xu_generalizable_2025} & ViT-L/16 & self-distillation, ${\sim}$100k MRI, CT, PET volumes & 1024 \\
    SAM-Med3D~\cite{wang_sam-med3d_2025} & ViT-B/16 & promptable segmentation, ${\sim}$22k CT, MRI volumes & 384 \\
    \bottomrule
  \end{tabular}
\end{table}

\subsection{Preprocessing}

Every volume, regardless of source or encoder, is reoriented to RAS,
resampled trilinearly to 2~mm isotropic spacing and centre-cropped or
zero-padded to $128^{3}$ voxels, a 256~mm box that contains a whole head
without clipping.\footnote{Five BrainMetShare scans have a header voxel size
(0.9375~mm) that disagrees with their affine (1.0~mm); the header value is
used.} Each encoder then receives only the resizing and intensity scaling
used in its pretraining, without additional skull stripping or registration
(Table~\ref{tab:encoder_input}). The Handcrafted statistics are computed
before intensity scaling, so that raw intensity scale remains among their
features. BrainIAC and BrainFM were pretrained on skull-stripped, registered
scans, whereas 10 of the 17 datasets are not skull-stripped; a control in
Section~\ref{sec:controls} tests whether this mismatch explains their
results.

\begin{table}[t]
  \centering
  \small
  \caption{Encoder-specific input size and intensity scaling, applied after
  the shared geometry pipeline.}
  \label{tab:encoder_input}
  \resizebox{\textwidth}{!}{%
  \begin{tabular}{lll}
    \toprule
    Encoder & Input size & Intensity scaling \\
    \midrule
    Untrained CNN, MedicalNet, SwinUNETR, SAM-Med3D & $128^{3}$ & $z$-score, non-zero voxels \\
    BrainIAC & $96^{3}$  & $z$-score, non-zero voxels \\
    3DINO    & $112^{3}$ & 0.05--99.95th percentile to $[-1,1]$ \\
    BrainFM  & $128^{3}$ & min--max to $[0,1]$ \\
    Handcrafted & $128^{3}$ & none \\
    \bottomrule
  \end{tabular}}
\end{table}

\section{Evaluation Protocol}
\label{sec:protocol}

Each representation is examined with respect to two labels: the source
dataset and the imaging sequence (T1, T1c, T2 or FLAIR), which serves as the
content label. PD occurs in a single dataset and is excluded. Features are
standardized and, where they have more than 100 dimensions, reduced to 100
principal components. All classifiers are $\ell_2$-regularized
logistic-regression probes with balanced class weights, implemented in
scikit-learn~\cite{pedregosa_scikit-learn_2011}, and data are split at the
patient level, so that scans of one patient never appear in both training
and test data.

\paragraph{Decodability.}
On the pooled corpus, the probe predicts the source dataset (17 classes) and
the sequence under five-fold cross-validation repeated three times; here
standardization and PCA are fitted once on the whole corpus, a step that
uses no labels. The sequence is also predicted within each of the ten
multi-sequence datasets (five-fold cross-validation, with standardization and
PCA fitted on the training folds). Performance is reported as balanced
accuracy and, for the pooled probes, compared against a null of 100 label
permutations. Per-scan acquisition properties derived from the images and
headers (template space, bias-field strength, a ghosting and motion index,
and, for IXI, the scanner) are probed in the same way.

\paragraph{Cross-dataset transfer.}
The sequence probe is trained on 16 datasets and evaluated on the held-out
one, with standardization and PCA refitted on the training datasets of each
split. This succeeds only if the held-out dataset encodes the sequence along
the same feature directions as the training datasets. The primary metric is
the balanced accuracy averaged over the ten multi-sequence held-out
datasets; plain accuracy on the seven single-sequence datasets is reported
separately. Representations are compared with a paired two-sided Wilcoxon
signed-rank test over held-out datasets.

\paragraph{Neighbourhood structure.}
Without training a classifier, the ten nearest neighbours of each volume
(cosine distance, excluding scans of the same patient) show whether the
feature space is organized locally by dataset or by sequence.

\paragraph{Direction consistency.}
We describe the score for the sequence; for sex it is defined in the same way
with the male--female direction. A linear probe trained on several datasets transfers to a new one only if the
new dataset separates the sequences along similar directions. We measure this
directly, inside each leave-one-dataset-out split and in the same feature space
as the probe. For a dataset $d$ with sequence set $M_d$, let $\mu_{d,m}$ be
the mean feature vector of sequence $m$, and let
$c_d=\frac{1}{|M_d|}\sum_{m\in M_d}\mu_{d,m}$ be the dataset's
class-balanced centre. The sequence offsets $o_{d,m}=\mu_{d,m}-c_d$ describe
how each sequence differs from the rest of its own dataset, independently of
where the dataset lies in feature space. The consistency of the held-out
dataset $h$ with a set of reference datasets $R$ is
\begin{equation}
  C(h;R)=\frac{1}{|M_h|}\sum_{m\in M_h}\cos\!\Big(o_{h,m},\;
  \frac{1}{|R_m|}\sum_{d\in R_m}o_{d,m}\Big),
\end{equation}
where $R_m\subseteq R$ are the reference datasets that contain sequence $m$
and only multi-sequence datasets are used. With $R$ the training datasets,
$C$ uses the held-out labels and therefore \emph{explains} transfer to that
dataset rather than predicting it. A version that needs no target data, the
\emph{pool consistency}, averages $C(d;R\setminus\{d\})$ over the training
datasets $d$ themselves; it characterizes a representation before it is
applied to a new dataset. The score is the analogue, for datasets and class
means, of the cross-lingual alignment measures that predict zero-shot
transfer in multilingual language models~\cite{muller_first_2021,
deshpande_when_2022}.

\paragraph{Centring tests.}
Two centring tests separate a shift of each dataset's features from a
disagreement of directions. \emph{Label-free centring} subtracts each
dataset's feature mean, including that of the held-out dataset, before the
probe is trained and applied; it can be used at test time. \emph{Class-balanced
centring} subtracts $c_d$ instead, which removes any shift without being
affected by the mix of sequences in each dataset, but uses the held-out labels
and is therefore an oracle. If transfer failed because each dataset is
displaced as a whole, the oracle would recover it; if it failed because
directions disagree, neither test would help.

\paragraph{Sex as a second content label.}
Sex is recorded for 2{,}186 volumes in ten datasets (Table~\ref{tab:datasets}:
3D-MR-MS, ISLES22, IXI, MS-60, MSLesSeg, NSK-GBM, OASIS-1, OASIS-2, UMF-PD and
UPENN-GBM). The same leave-one-dataset-out protocol is applied to these ten
datasets, with a balanced-class probe for sex. Because several datasets
contain more than one sequence, the male--female direction of a dataset is
computed inside each sequence with at least three scans per sex and then
averaged, so that a different mix of sequences between the sexes does not
enter it; consistency is the cosine between this direction and the mean
direction of the training datasets, and the centring tests are defined as
above.

\paragraph{Post-hoc removal of the dataset signal.}
Four standard methods are applied unmodified, covering linear erasure,
distribution alignment and non-linear adversarial training.
INLP~\cite{ravfogel_null_2020} repeatedly projects the features onto the
nullspace of a linear classifier, first for the per-scan acquisition
properties and then for the source dataset; a step is discarded if it lowers
sequence decodability by more than five points, a threshold fixed in
advance. LEACE~\cite{belrose_leace_2023} removes, in closed form, the linear
subspace that predicts the source dataset while changing the features as
little as possible (rank 15, one less than the number of training datasets).
CORAL~\cite{sun_deep_2016} aligns rather than erases, recolouring each
dataset's features to the mean and covariance of the training datasets; it
also uses the unlabelled features of the held-out dataset. The adversarial
method~\cite{ganin_domain-adversarial_2016, dinsdale_deep_2021} trains a
small network, through a gradient-reversal layer, to map the features so
that a dataset classifier fails. Each method has a control that should not
help: a random projection of the same rank (INLP, LEACE), alignment to a
wrong dataset (CORAL), and an untrained network of the same shape
(adversarial). Methods and controls are fitted on the 16 training datasets of
every leave-one-dataset-out split, after which the sequence probe is trained
and tested exactly as before, so the values before removal equal those of the
transfer evaluation. Dataset and sequence decodability are also measured
inside the training datasets before and after removal, to check what each
method actually removed. For sex, LEACE and CORAL with their controls are
applied in the same way; INLP and the adversarial network are not, because
their stopping rule and training objective here are built to retain the
sequence.

\paragraph{Predicting failure without target labels.}
A user deploying a frozen encoder on a new dataset usually has no labels for
it. We therefore compute consistency from the unlabelled held-out features:
they are clustered with $k$-means, with $k$ the number of sequences in the
dataset (known from file names in practice), and the cluster offsets are
matched to the training sequence offsets by maximizing the summed cosine
(Hungarian algorithm); the score is the mean matched cosine. It is compared
with two standard label-free estimators of accuracy under distribution shift:
the average maximum softmax probability of the
probe~\cite{hendrycks_baseline_2017}, and average thresholded confidence
(ATC)~\cite{garg_leveraging_2022}, whose threshold is set by patient-grouped
cross-validation on the training datasets. Each score is evaluated by its
Spearman correlation with the measured transfer across the held-out datasets
of one representation, the question a user faces, and averaged over
representations.

\section{Results}
\label{sec:results}

\begin{table}[t]
\centering
\small
\caption{Properties encoded by each frozen representation.
\emph{Decodable}: balanced accuracy of patient-level cross-validation for the
source dataset and the sequence on the pooled corpus (chance 0.06 and 0.25;
all outside the permutation null, $p<0.01$), and for the sequence within each
multi-sequence dataset (mean over ten datasets). \emph{Across datasets}:
leave-one-dataset-out sequence prediction, as balanced accuracy averaged over
the ten multi-sequence held-out datasets and as accuracy averaged over the
seven single-sequence datasets.}
\label{tab:main}
\resizebox{\textwidth}{!}{%
\begin{tabular}{l l c c c c c}
\toprule
 & & \multicolumn{3}{c}{\textbf{Decodable}} & \multicolumn{2}{c}{\textbf{Across datasets}} \\
\cmidrule(lr){3-5}\cmidrule(lr){6-7}
\textbf{Group} & \textbf{Representation} & \textbf{Dataset} & \textbf{Sequence} & \textbf{Seq.\ within} & \textbf{10 multi-seq.} & \textbf{7 single-seq.} \\
\midrule
Baseline encoders  & Handcrafted              & 0.92 & 0.72 & 0.85 & 0.44 & 0.46 \\
                   & Untrained CNN            & 0.94 & 0.88 & 0.93 & 0.49 & 0.73 \\
                   & MedicalNet               & 0.94 & 0.87 & 0.89 & 0.44 & 0.58 \\
                   & SwinUNETR                & 0.91 & 0.83 & 0.91 & 0.47 & 0.67 \\
\midrule
Brain-specific FMs & BrainIAC                 & 0.84 & 0.76 & 0.84 & 0.50 & 0.53 \\
                   & BrainFM                  & 0.93 & 0.89 & 0.93 & 0.55 & 0.75 \\
\midrule
Broad medical FMs  & 3DINO                    & 0.95 & 0.95 & 0.97 & \textbf{0.72} & \textbf{0.97} \\
                   & SAM-Med3D                & 0.94 & 0.92 & 0.93 & 0.69 & 0.85 \\
\midrule
Controls           & 3DINO, random weights    & 0.79 & 0.76 & 0.86 & 0.50 & 0.67 \\
                   & 3DINO, $z$-score input   & 0.94 & 0.92 & 0.94 & 0.62 & 0.78 \\
                   & MedicalNet, percentile input & 0.94 & 0.90 & 0.93 & 0.60 & 0.67 \\
\bottomrule
\end{tabular}%
}
\end{table}

\subsection{The sequence is encoded within every dataset but lost across datasets}
\label{sec:transfer}

\begin{figure}[t]
\centering
\includegraphics[width=0.9\textwidth]{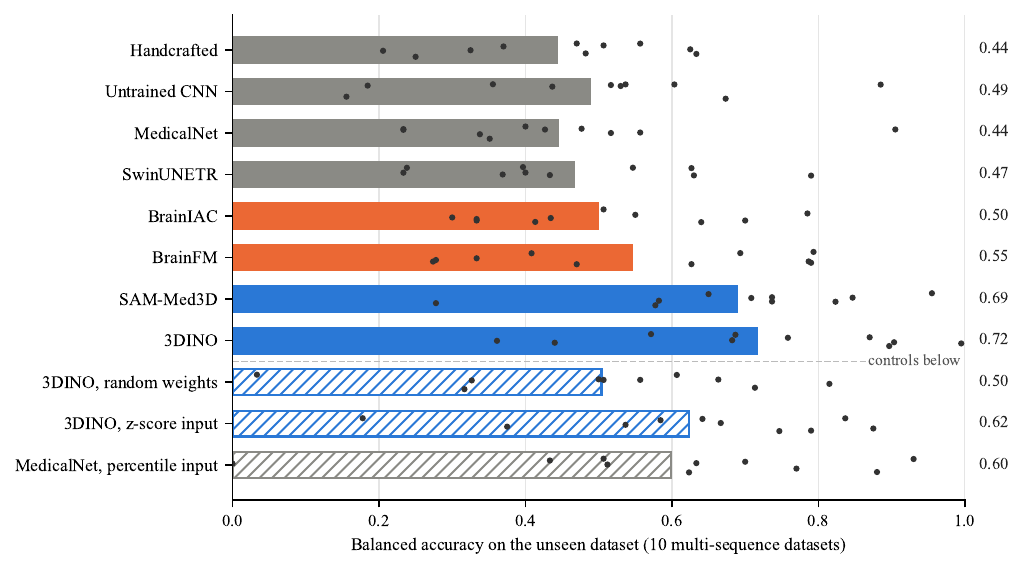}
\caption{Leave-one-dataset-out transfer of the imaging sequence. Balanced
accuracy on each of the ten held-out datasets with more than one sequence
(dots) and their mean (bar and number). Grey: baseline encoders; orange:
brain-specific foundation models; blue: broadly pretrained medical foundation
models; hatched: controls.}
\label{fig:transfer}
\end{figure}

Within a single dataset, every representation distinguishes the sequences
reliably (0.84--0.97, Table~\ref{tab:main}), and the sequence is also
decodable on the pooled corpus (0.72--0.95), where the probe sees every
dataset during training. Yet a probe trained on 16 datasets largely fails on
the seventeenth (Figure~\ref{fig:transfer}). For the baseline encoders and
both brain-specific models, balanced accuracy falls to 0.44--0.55. The two
broadly pretrained models, 3DINO and SAM-Med3D, keep more (0.72 and 0.69),
but still lose 0.24--0.25 of their within-dataset accuracy, too much for a
sequence classifier to be reliable on a new dataset without adaptation. Sex
fails in the same way: across datasets, every representation stays between
0.51 and 0.69 balanced accuracy against a chance level of 0.50
(Table~\ref{tab:sex}).

\subsection{Why transfer fails}
\label{sec:why}

Since the sequence is present in every dataset, the failure is not missing
information. We test three explanations in turn: the features encode the
source dataset, each dataset is shifted as a whole, or the directions that
separate the classes differ between datasets. We then apply the same tests to
sex.

\subsubsection{Encoding the source dataset does not explain the failure}

Every representation identifies the source dataset with a balanced accuracy
of 0.84--0.95, against a chance level of 0.06 (Table~\ref{tab:main};
all $p<0.01$). Most of this signal requires no learning: the Untrained CNN
matches MedicalNet (both 0.94), and the Handcrafted statistics reach 0.92.
The randomly initialized ViT-L is the exception, reaching 0.79 before and
0.95 after 3DINO pretraining. Per-scan acquisition properties (template
space, bias field, ghosting and motion, and IXI scanner) are likewise
decodable (0.85--1.00, chance 0.25--0.33). This agrees with earlier reports
that the source study can be recovered from brain
MRI~\cite{wachinger_detect_2021, glocker_machine_2019} and from natural
images~\cite{torralba_unbiased_2011}, even with randomly initialized
features~\cite{rahbar_frozen_2026}.

\begin{figure}[t]
\centering
\includegraphics[width=0.6\textwidth]{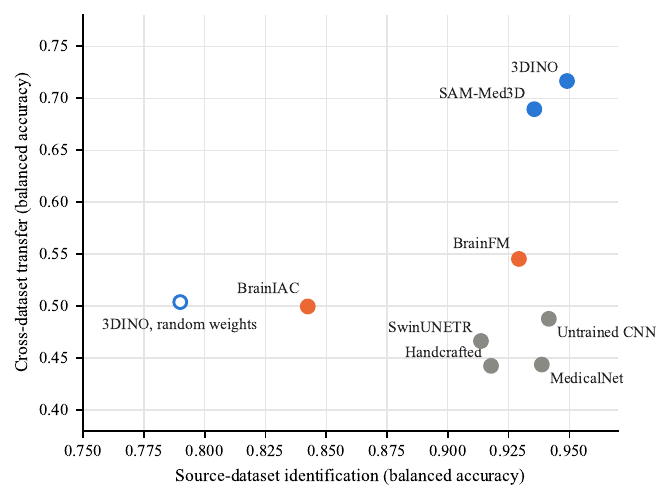}
\caption{Source-dataset decodability (balanced accuracy, 17 classes) against
cross-dataset sequence transfer (leave-one-dataset-out balanced accuracy on
the ten multi-sequence held-out datasets). Grey: baseline encoders; orange:
brain-specific foundation models; blue: broadly pretrained medical foundation
models; open circle: 3DINO with random weights.}
\label{fig:identify}
\end{figure}

Encoding the source dataset, however, does not go together with poor
transfer (Figure~\ref{fig:identify}). 3DINO separates the source datasets
best and also transfers the sequence best; the baseline encoders separate
datasets almost as well but transfer far worse; and the random-weight ViT-L
separates datasets least yet transfers poorly. Both kinds of information can
coexist in the same features, in line with observations in
mammography~\cite{nguyen_benchmarking_2026}, so dataset decodability is not
the explanation.

The neighbourhood structure reflects the same difference. This comparison of
same-dataset and same-sequence neighbours follows the robustness index of
PathoROB~\cite{komen_towards_2025}, which compares same-centre and
same-biology neighbours in pathology. Averaged over the ten multi-sequence
datasets, most representations have more neighbours from the same dataset
than with the same sequence (e.g.\ MedicalNet 0.86 against 0.77, BrainFM 0.88
against 0.77); locally, they are organized primarily by dataset. 3DINO is the
only representation in which the same sequence is at least as common as the
same dataset (0.89 against 0.87), despite separating datasets best.

\subsubsection{The failure is not a shift of each dataset}

If each dataset were displaced as a whole in feature space, removing each
dataset's mean would recover transfer. It does not
(Figure~\ref{fig:consistency}b). Neither centring test changes sequence
transfer significantly for any representation ($p\ge0.16$); most
representations lose slightly, and 3DINO gains 0.06 without reaching
significance. Label-free and class-balanced centring agree to within 0.01,
so the different mix of sequences in each dataset is not the cause either.

\subsubsection{The failure is a disagreement of sequence directions}
\label{sec:consistency}

\begin{figure}[t]
\centering
\includegraphics[width=\textwidth]{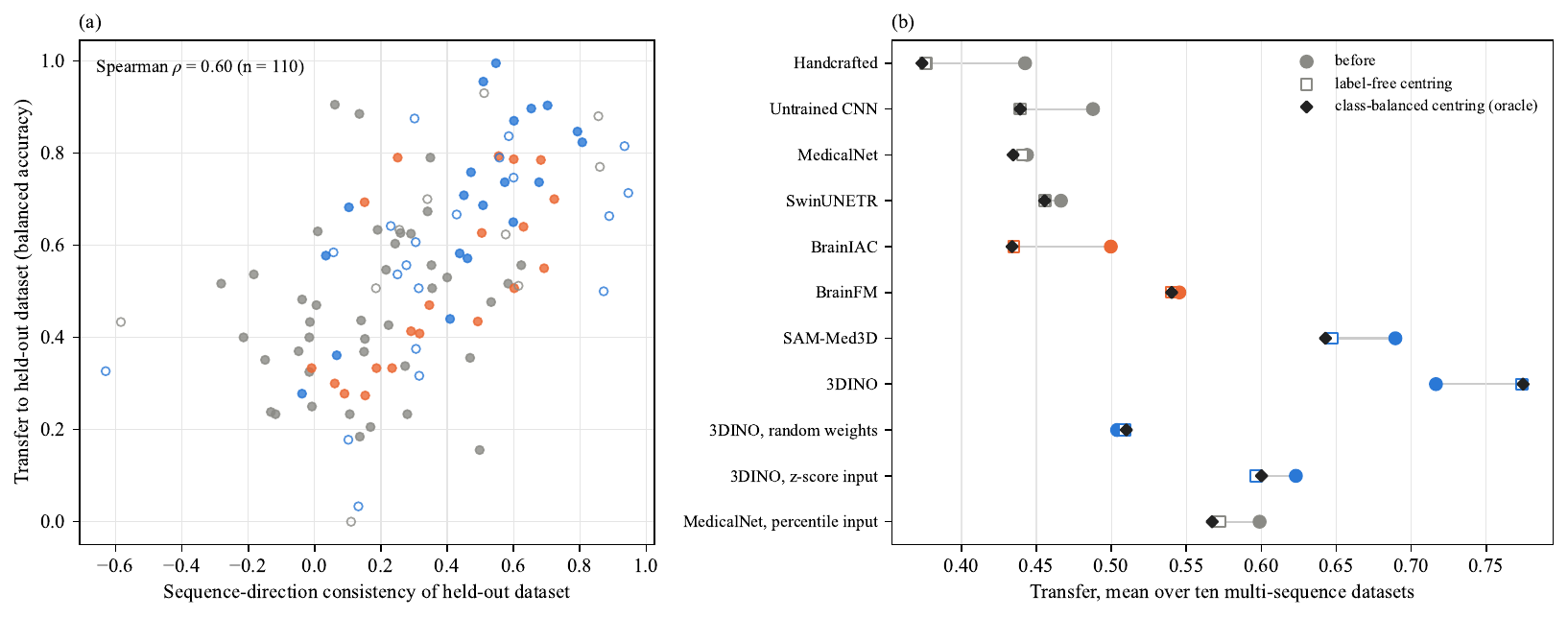}
\caption{Sequence-direction consistency and transfer. (a) Consistency of each
held-out multi-sequence dataset with the training datasets against transfer
to that dataset, for every representation (filled) and control (open). (b)
Mean transfer before and after centring each dataset's features, label-free
(squares) or class-balanced using the held-out labels (oracle, diamonds).
Grey: baseline encoders; orange: brain-specific foundation models; blue:
broadly pretrained medical foundation models.}
\label{fig:consistency}
\end{figure}

Direction consistency tracks transfer. Across all representations and held-out datasets, a held-out
dataset whose sequence offsets agree with those of the training datasets is
one to which the sequence transfers (Spearman $\rho=0.60$, $p<10^{-10}$,
$n=110$; Figure~\ref{fig:consistency}a). The relation also holds within
representations, positive for 10 of 11 with a mean $\rho$ of 0.57, so it
identifies which datasets a given representation fails on and is not only a
difference between representations. The baseline encoders have the least
consistent directions (0.11--0.21), BrainFM reaches 0.30, and 3DINO and
SAM-Med3D reach 0.45 and 0.48; BrainIAC (0.46) is discussed below. The input-scaling controls move
consistency and transfer together: percentile scaling raises MedicalNet's
consistency from 0.21 to 0.37, and $z$-score input lowers 3DINO's from 0.45
to 0.34.

Because this score uses the held-out labels, part of this relation is
expected by construction. Two versions without them show it is more than
that. Measured on the training datasets alone, the pool consistency ranks the 11 representations and controls by their transfer to
unseen datasets ($\rho=0.69$, $p=0.02$), without any data from the target.
Computed from clusters of the unlabelled held-out dataset, it ranks the
held-out datasets of a representation by transfer with a mean $\rho$ of 0.37,
better than average confidence in 10 of 11 representations and than ATC in all
11, which barely track transfer within a representation ($\rho=0.03$ and 0.01;
Appendix Table~\ref{tab:failure}). It works best where the sequences form
clear clusters, as for 3DINO and SAM-Med3D ($\rho=0.79$ and 0.92), whereas
confidence remains the better way to compare whole representations ($\rho=0.91$
against 0.11).

Consistency is a necessary rather than a sufficient condition: BrainIAC
(0.46) and 3DINO with random weights (0.44) have consistent directions yet
transfer poorly. Both also have among the lowest within-dataset sequence
accuracies (0.84 and 0.86, Table~\ref{tab:main}): their sequence offsets
point the same way across datasets but are small relative to the spread
within each sequence.

\subsubsection{Sex: directions disagree, and part of the failure is a shift}
\label{sec:sex}

Direction consistency again explains which held-out datasets fail,
more weakly than for the sequence (Spearman $\rho=0.42$, $p<10^{-5}$,
$n=110$; positive within 10 of 11 representations, mean $\rho=0.33$), and the
mean consistency of a representation tracks its mean transfer ($\rho=0.71$,
$p=0.015$, $n=11$). The pool consistency, computed without the target,
points the same way but is not significant here ($\rho=0.59$, $p=0.06$).

Unlike for the sequence, centring helps in part. Label-free centring raises
the sex transfer of the Handcrafted statistics from 0.51 to 0.61 ($p=0.004$)
and moves MedicalNet and BrainFM up by 0.03--0.06 without reaching
significance, whereas it changes the broadly pretrained models by at most
0.02 (Table~\ref{tab:sex}). For these representations part of the failure is
therefore a shift of each dataset as a whole, which the centring test
separates from a disagreement of directions.

\begin{table}[t]
\centering
\small
\caption{Sex as a second content label. Leave-one-dataset-out balanced
accuracy averaged over the ten held-out datasets with sex labels; consistency
of the held-out male--female direction with the training datasets. Control in
brackets: a random projection of the same rank (LEACE) or alignment to a wrong
dataset (CORAL). $^{*}$: differs from before, $p<0.05$, paired Wilcoxon test.}
\label{tab:sex}
\begin{tabular}{l c c c c c}
\toprule
\textbf{Representation} & \textbf{Consistency} & \textbf{Before} & \textbf{Centring} & \textbf{LEACE} & \textbf{CORAL} \\
\midrule
Handcrafted   & 0.34 & 0.51 & 0.61$^{*}$ & 0.52 (0.51) & 0.65$^{*}$ (0.52) \\
Untrained CNN & 0.65 & 0.64 & 0.64 & 0.63 (0.65) & 0.70 (0.59) \\
MedicalNet    & 0.49 & 0.65 & 0.70 & 0.62$^{*}$ (0.62) & 0.67 (0.58) \\
SwinUNETR     & $-$0.05 & 0.60 & 0.60 & 0.60 (0.60) & 0.63 (0.60) \\
\midrule
BrainIAC      & 0.13 & 0.54 & 0.54 & 0.54 (0.55) & 0.53 (0.53) \\
BrainFM       & 0.40 & 0.65 & 0.68 & 0.65 (0.64) & 0.69 (0.57) \\
\midrule
3DINO         & 0.43 & 0.65 & 0.67 & 0.65 (0.63) & 0.65 (0.58) \\
SAM-Med3D     & 0.60 & 0.69 & 0.69 & 0.68 (0.66) & 0.68 (0.61) \\
\bottomrule
\end{tabular}
\end{table}

\subsection{Why removing the dataset signal does not help}
\label{sec:removal}

The removal methods succeed at what
they are designed to do. For the seven network representations, LEACE lowers
dataset decodability inside the training datasets from 0.84--0.95 to
0.03--0.26 and CORAL to at most 0.11, while INLP (0.67--0.91) and the
adversarial network (0.72--0.90) remove less (Appendix
Table~\ref{tab:removal_decode}). Yet none of the 32 method--representation
combinations improves sequence transfer significantly
(Table~\ref{tab:removal}); almost all lower it, and seven do so
significantly, mostly after LEACE and the adversarial network. The
single-sequence datasets show the same pattern, with every method lowering
accuracy for every network representation. Removing the dataset directions is
also no better than removing random ones: for no network representation do
INLP or LEACE significantly beat their random-projection controls, and CORAL,
although better than aligning to a wrong dataset, still lowers transfer.
This extends the observation that INLP and ComBat bring no significant gain
on held-out sites of one study~\cite{rahbar_frozen_2026} to independent
datasets, two further methods, matched controls for each method and eight
representations. It also rules out the explanation offered there, that the
encoder output carried too little clinical information to begin with: here
the sequence is strongly encoded, and removal still fails.

The reason is that the dataset and the sequence are not stored separately.
When the difference between sequences is itself different in each dataset,
the directions that tell datasets apart include these per-dataset sequence
differences. Removing the dataset signal therefore also removes sequence
information: inside the training datasets, sequence decodability falls by
0.14--0.24 after LEACE and by 0.26--0.28 after CORAL for every network
representation. This loss is specific to the dataset directions: removing 15
random directions, the LEACE control, costs only 0.01--0.05, and the
difference holds for every representation ($p<0.001$, paired over the 17
splits; Appendix Table~\ref{tab:removal_decode}). Alignment fails for the reason found in the centring tests:
matching the first two moments of each dataset cannot rotate inconsistent
sequence directions into agreement. 3DINO is a partial exception: there,
INLP removes 16 directions on average, against 3--9 for the other
representations, lowering dataset decodability from 0.95 to 0.67 while
sequence decodability falls only from 0.95 to 0.90. The features that encode
the sequence most consistently are thus also those in which the dataset
signal is most separable from it. After any correction, no representation
other than 3DINO and SAM-Med3D exceeds 0.57, whereas the uncorrected 3DINO
features reach 0.72: how consistently the sequence is encoded is set when the
features are learned.

For sex, where part of the failure is a shift (Section~\ref{sec:sex}),
alignment can help. CORAL, which includes the per-dataset centring, raises
the Handcrafted statistics from 0.51 to 0.65 ($p=0.002$), while LEACE never
helps and lowers MedicalNet ($p=0.04$; Table~\ref{tab:sex}). Alignment thus
helps where the failure is a shift but not where directions disagree, and the
centring test tells the two apart.

\begin{table}[t]
\centering
\small
\caption{Sequence transfer before and after removing the dataset signal.
Leave-one-dataset-out balanced accuracy averaged over the ten multi-sequence
held-out datasets; methods fitted on the training datasets of each split.
Control in brackets: a random projection of the same rank (INLP, LEACE),
alignment to a wrong dataset (CORAL), an untrained network of the same shape
(adversarial). $^{*}$: after differs from before, $p<0.05$, paired Wilcoxon
test over the ten datasets.}
\label{tab:removal}
\begin{tabular}{l c c c c c}
\toprule
\textbf{Representation} & \textbf{Before} & \textbf{INLP} & \textbf{LEACE} & \textbf{CORAL} & \textbf{Adversarial} \\
\midrule
Handcrafted   & 0.44 & 0.41 (0.44) & 0.42 (0.40) & 0.37 (0.27) & 0.42 (0.43) \\
Untrained CNN & 0.49 & 0.48 (0.47) & 0.41$^{*}$ (0.44) & 0.42 (0.28) & 0.45 (0.44) \\
MedicalNet    & 0.44 & 0.44 (0.44) & 0.43 (0.42) & 0.42 (0.27) & 0.40 (0.45) \\
SwinUNETR     & 0.47 & 0.41$^{*}$ (0.45) & 0.44 (0.40) & 0.45 (0.30) & 0.36 (0.41) \\
\midrule
BrainIAC      & 0.50 & 0.48 (0.50) & 0.44$^{*}$ (0.48) & 0.43 (0.25) & 0.40$^{*}$ (0.45) \\
BrainFM       & 0.55 & 0.57 (0.55) & 0.54 (0.53) & 0.48 (0.33) & 0.50 (0.51) \\
\midrule
3DINO         & 0.72 & 0.72 (0.73) & 0.70 (0.73) & 0.66 (0.32) & 0.51$^{*}$ (0.69) \\
SAM-Med3D     & 0.69 & 0.67 (0.68) & 0.63 (0.64) & 0.57$^{*}$ (0.31) & 0.54$^{*}$ (0.56) \\
\bottomrule
\end{tabular}
\end{table}

\subsection{Which representations encode the sequence consistently}
\label{sec:models}

The two broadly pretrained models are the only ones with sequence directions
consistent enough to transfer. 3DINO and SAM-Med3D exceed the Untrained CNN,
which reflects the architecture alone, on every one of the ten held-out
datasets ($+0.23$ and $+0.20$, $p=0.002$ each), so the gain is not driven by
a few favourable datasets. Pretraining on brain MRI does not have this
effect: BrainIAC and BrainFM do not transfer significantly better than the
Untrained CNN ($+0.01$, $p=0.85$; $+0.06$, $p=0.13$), so matching the
pretraining domain to the target anatomy is not sufficient. Because two
different objectives, self-distillation and promptable segmentation, reach a
similar level ($p=0.24$), the breadth of the pretraining data likely matters
more than the objective. The gain is specific to the sequence: for sex, 3DINO
transfers no better than the other network representations (0.65, against
0.54--0.65) and SAM-Med3D only slightly better (0.69). Broad pretraining
appears to align strong, global properties such as contrast across datasets,
but not necessarily subtle anatomical ones.

These results refine two earlier observations. Rahbar~\cite{rahbar_frozen_2026}
found no reliable advantage of brain-pretrained or CT-pretrained SwinUNETR
encoders over an untrained copy; our CT-pretrained SwinUNETR likewise does not
beat the Untrained CNN, so pretraining as such is not enough, but broad
pretraining is. Mielcarz et al.~\cite{mielcarz_3d_2026} found 3DINO the most
stable, BrainFM intermediate and BrainIAC the most sensitive under simulated
artifacts; the three models fall in the same order here under real shift
between independent datasets.

\paragraph{Controls.}
\label{sec:controls}
Four controls rule out simpler explanations for these differences.
First, the gain is not due to the architecture: the ViT-L of 3DINO with
random weights reaches only 0.50, like the baseline encoders and
significantly below the pretrained model ($p=0.01$), so the consistency is
acquired during pretraining. Second, input normalization explains part of the
gain but not all of it. Replacing 3DINO's percentile scaling with the shared
$z$-score input lowers its transfer to 0.62 ($p=0.01$), and giving MedicalNet
percentile scaling raises it from 0.44 to 0.60 ($p=0.03$); yet 3DINO with
$z$-score input remains above the Untrained CNN ($+0.14$, $p=0.004$), and
MedicalNet with percentile input remains below both broadly pretrained models
($p\le0.03$). Third, the brain-specific models do not fail merely because
half of the held-out datasets retain the skull: on the five skull-stripped
multi-sequence datasets, where the input matches their pretraining, BrainIAC
and BrainFM reach 0.55 and 0.64, against 0.81 for 3DINO and 0.75 for
SAM-Med3D. Finally, the advantage is not an artifact of pretraining overlap.
UPENN-GBM, OASIS-1 and OASIS-2 may be part of the pretraining data of 3DINO
or SAM-Med3D; excluding UPENN-GBM, the only one of the three with more than
one sequence, leaves both above the Untrained CNN ($+0.25$ and $+0.20$,
$p=0.004$ each, $n=9$).

\section{Discussion}

\paragraph{Why frozen features fail across datasets.}
Every representation tested identifies the source dataset and recognizes the
sequence within each dataset, yet neither predicts transfer. The failure is
not missing information, dataset encoding or a shift of each dataset; the
directions that separate the classes differ between datasets. Dataset
decodability and within-dataset or pooled accuracy are therefore poor guides
to whether a representation is fit for pooling, and consistency must be
measured on held-out datasets. The same account explains why removal fails:
the directions that tell datasets apart include the per-dataset sequence
differences, so erasing or aligning the dataset signal removes content too.
This agrees with the observation that removing a site subspace can harm
segmentation because site and anatomy share
directions~\cite{rahbar_frozen_2026}. Correction has succeeded where \emph{paired}
scans of the same subjects are available~\cite{donle_featmap_2026}, but public
datasets are independent populations with only a dataset label; there,
adapting the encoder with labelled target data remains the practical option.

\paragraph{Direction consistency as a diagnostic.}
Direction consistency is computed from class means alone and supports three
practical checks. Before deployment, the
pool consistency of the training datasets indicates which representation
will transfer, significantly for the sequence and in the same direction for
sex. On a new unlabelled dataset, cluster consistency warns which datasets a
chosen encoder will fail on, a task on which standard confidence-based
estimators~\cite{hendrycks_baseline_2017, garg_leveraging_2022} fail under
this kind of shift. And the centring test, run on labelled datasets, shows
whether aligning datasets afterwards is worth trying at all. If a few target labels are available, the
probe's accuracy on them is the more direct check. Consistency is necessary
rather than sufficient, since the class offsets must also be large relative
to the spread within each class, and the label-free variant is modest in
absolute terms; a score that combines direction and separation is a natural
next step.

\paragraph{Which features transfer: breadth of pretraining, not domain match.}
The two models pretrained on broad multi-organ CT and MRI collections encode
the sequence consistently; the two brain-specific models, pretrained on the
target anatomy, do not do significantly better than a network with random
weights. The controls attribute the advantage to pretraining rather than to
architecture, input scaling, skull stripping or data overlap. A plausible
reading is that exposure to many scanners and protocols forces contrast to be
encoded in shared terms, whereas a narrower brain-only corpus does not,
although with two models per group this remains a hypothesis. It is
consistent with the finding in mammography that domain-specific pretraining
does not guarantee robustness to dataset shift~\cite{nguyen_benchmarking_2026},
with the robustness of 3DINO to simulated
artifacts~\cite{mielcarz_3d_2026} and with the poor cross-scanner reliability
of BrainIAC~\cite{navarro-gonzalez_cross-scanner_2026}, and it shows that the
absence of a pretraining advantage reported for SwinUNETR
encoders~\cite{rahbar_frozen_2026} does not hold for every model. Encoders
supervised with acquisition or anatomical
metadata~\cite{navarro-gonzalez_cross-scanner_2026, avci_metadata_2026} are a
natural next comparison. The advantage was small for sex, a subtler
property: only SAM-Med3D was slightly ahead.

\paragraph{Limitations.}
The sequence is the only content label shared by all 17 datasets; it is a
global, comparatively easy property. Sex extends the analysis to an
anatomical label, but in ten datasets only, with weaker effects, and with
LEACE and CORAL as the only removal methods. The label-free failure
prediction assumes that the number of sequences in a new dataset is known and
reaches only a moderate correlation. Transfer is measured with linear
probes on frozen features; non-linear probes or fine-tuning may behave
differently. The paired tests use ten multi-sequence held-out datasets, which
limits statistical power; the pooled consistency correlation counts each
dataset once per representation and is therefore not a test on independent
samples, and the ranking of representations by pool consistency rests on 11
points. Each encoder receives the input scaling of its own
pretraining, so part of the differences between models may reflect
preprocessing, although the scaling controls bound this effect. The removal
methods were applied with standard settings and a dataset label only; methods
that use paired data or richer supervision were outside the scope. Finally,
pretraining data of the foundation models are not fully documented, so the
overlap control can only exclude datasets known or likely to be included.

\section{Conclusion}

Across 17 public brain MRI datasets, frozen features fail on new datasets
not because the content is missing, nor because they encode the source
dataset or because each dataset is shifted, but because the directions that
separate the classes differ between datasets. This direction consistency
tracks transfer for the sequence and for sex, ranks representations from the training datasets alone
and, computed on an unlabelled new dataset, warns of failure better than
confidence-based estimators. The same account explains why four standard
methods for removing the dataset signal erased it without improving sequence
transfer, and why alignment helped only where a centring test showed that the
failure was a shift. Only two broadly pretrained medical foundation models
encoded the sequence consistently enough to transfer; brain-specific
pretraining did not. For anyone pooling public data, consistency across
datasets should be treated as a property of how features are learned and
verified on held-out datasets, not something to be corrected afterwards.

\section*{Code and Data Availability}

Code, the list of sampled scans and the result tables are available at
\url{https://github.com/BrainFM/brainfm-profiler/tree/main/representation_utility}. All 17 datasets are public
and can be obtained from their original sources under their own licences.
Because several of these licences do not permit redistribution of the data
or of material derived from it, we do not redistribute the images or the
extracted embeddings; the embeddings can be regenerated with the released
code from the listed scans.

\bibliographystyle{unsrtnat}
\bibliography{references}

\appendix

\section{Decodability After Removal}

\begin{table}[H]
\centering
\small
\caption{Dataset and sequence decodability inside the training datasets,
before and after each removal method, averaged over the
leave-one-dataset-out splits (balanced accuracy). Chance: 0.06 for the
dataset and 0.25 for the sequence. Adv.: adversarial. Rand.: removing 15
random directions, the control for LEACE.}
\label{tab:removal_decode}
\resizebox{\textwidth}{!}{%
\begin{tabular}{l cc cccc ccccc}
\toprule
 & \multicolumn{2}{c}{\textbf{Before}} & \multicolumn{4}{c}{\textbf{Dataset after}} & \multicolumn{5}{c}{\textbf{Sequence after}} \\
\cmidrule(lr){2-3}\cmidrule(lr){4-7}\cmidrule(lr){8-12}
\textbf{Representation} & Dataset & Seq. & INLP & LEACE & CORAL & Adv. & INLP & LEACE & Rand. & CORAL & Adv. \\
\midrule
Handcrafted   & 0.92 & 0.73 & 0.88 & 0.84 & 0.14 & 0.88 & 0.68 & 0.56 & 0.67 & 0.54 & 0.71 \\
Untrained CNN & 0.94 & 0.88 & 0.89 & 0.26 & 0.11 & 0.90 & 0.83 & 0.71 & 0.87 & 0.62 & 0.78 \\
MedicalNet    & 0.94 & 0.86 & 0.91 & 0.10 & 0.02 & 0.87 & 0.80 & 0.67 & 0.81 & 0.58 & 0.70 \\
SwinUNETR     & 0.91 & 0.83 & 0.87 & 0.10 & 0.03 & 0.85 & 0.78 & 0.59 & 0.79 & 0.56 & 0.69 \\
\midrule
BrainIAC      & 0.84 & 0.76 & 0.75 & 0.03 & 0.01 & 0.72 & 0.71 & 0.57 & 0.72 & 0.49 & 0.59 \\
BrainFM       & 0.93 & 0.89 & 0.90 & 0.23 & 0.07 & 0.90 & 0.84 & 0.68 & 0.88 & 0.62 & 0.80 \\
\midrule
3DINO         & 0.95 & 0.95 & 0.67 & 0.10 & 0.01 & 0.86 & 0.90 & 0.81 & 0.94 & 0.69 & 0.74 \\
SAM-Med3D     & 0.93 & 0.92 & 0.87 & 0.05 & 0.01 & 0.84 & 0.87 & 0.74 & 0.88 & 0.64 & 0.75 \\
\bottomrule
\end{tabular}}
\end{table}

\section{Label-Free Prediction of Transfer}

\begin{table}[H]
\centering
\small
\caption{Label-free prediction of sequence transfer to a held-out dataset.
Spearman correlation between each score and the measured balanced accuracy,
across the ten held-out datasets within each representation (mean over 11
representations and controls), across all 110 representation--dataset pairs,
and across the 11 representation means. ATC also misestimates the balanced
accuracy by 0.25 on average.}
\label{tab:failure}
\begin{tabular}{l c c c}
\toprule
\textbf{Score} & \textbf{Within representation} & \textbf{Pooled} & \textbf{Across representations} \\
\midrule
Average confidence~\cite{hendrycks_baseline_2017} & 0.03 & 0.26 & \textbf{0.91} \\
ATC~\cite{garg_leveraging_2022}                   & 0.01 & 0.11 & 0.68 \\
Cluster consistency (ours)                        & \textbf{0.37} & \textbf{0.27} & 0.11 \\
\bottomrule
\end{tabular}
\end{table}

\end{document}